\documentclass{article}

\usepackage{arxiv}

\usepackage[utf8]{inputenc} 
\usepackage[T1]{fontenc}    
\usepackage{hyperref}       
\usepackage{url}            
\usepackage{booktabs}       
\usepackage{amsfonts}       
\usepackage{nicefrac}       
\usepackage{microtype}      
\usepackage{lipsum}
\usepackage{graphicx}
\graphicspath{ {./images/} }

\title{FlowScope: Enhancing Decision Making by Time Series Forecasting based on Prediction Optimization using HybridFlow Forecast Framework}

\author{
Nitin Sagar Boyeena \\
    \textit{Dept. AI and Data Science} \\
    B V Raju Institute of Technology \\
    Telangana, India \\
    Email: nitinsagar2004@gmail.com,
\And
Begari Susheel Kumar \\
\textit{Dept. AI and Data Science} \\
    B V Raju Institute of Technology \\
    Telangana, India \\
    Email: specialsusheel@gmail.com
}

\begin{document}
\maketitle
\begin{abstract}
Time series forecasting is crucial in several sectors, such as meteorology, retail, healthcare, and finance. Accurately forecasting future trends and patterns is crucial for strategic planning and making well-informed decisions. In this case, it is crucial to include many forecasting methodologies. The strengths of Auto-regressive Integrated Moving Average (ARIMA) for linear time series, Seasonal ARIMA models (SARIMA) for seasonal time series, Exponential Smoothing State Space Models (ETS) for handling errors and trends, and Long Short-Term Memory (LSTM) Neural Network model for complex pattern recognition have been combined to create a comprehensive framework called FlowScope. SARIMA excels in capturing seasonal variations, whereas ARIMA ensures effective handling of linear time series. ETS models excel in capturing trends and correcting errors, whereas LSTM networks excel in reflecting intricate temporal connections. By combining these methods from both machine learning and deep learning, we propose a deep-hybrid learning approach FlowScope which offers a versatile and robust platform for predicting time series data. This empowers enterprises to make informed decisions and optimize long-term strategies for maximum performance.
\end{abstract}


\section{Introduction}
In the rapidly changing environment of contemporary business, where choices are increasingly guided by data, the skill of effectively predicting future trends has become of utmost importance. Time series forecasting involves analyzing and predicting successive data points, which provides significant insights into patterns, trends, and prospective consequences. In this context, the creation of FlowScope is a noteworthy achievement. It is a complete framework that effectively combines conventional and advanced forecasting methods to improve decision-making and maximize predictive precision.

To begin on the road of time series forecasting, it's necessary to comprehend the essence of time series data itself. At its heart, time series data consists of observations gathered at regular intervals across time, each point representing a snapshot of a specific variable's value. These data sets generally display various features such as trend, seasonality, cyclicality, and irregular fluctuations, each of which poses new problems and possibilities for predictive modeling.

\begin{figure}[htbp] 
    \centering
    \includegraphics[height=135 px, width = 250 px]{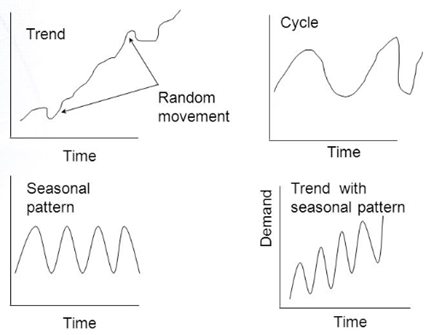} 
    \caption{Components of Time Series data}
\end{figure}

Consider, for instance, a business recording monthly sales numbers over many years. Such data may indicate a clear rising trend owing to general company growth, seasonality deriving from holiday purchasing peaks, and potentially irregular variations induced by external events such as economic downturns or supply chain interruptions. Understanding and accurately modeling these patterns are vital for accurate forecasting and informed decision-making.

Historically, classical statistical models have served as the backbone of time series forecasting. Among them, the Auto Regressive Integrated Moving Average (ARIMA) and Seasonal ARIMA (SARIMA) models stand out as cornerstones of analysis. ARIMA models, defined by their capacity to capture autoregressive, integrated, and moving average components, are well-suited for stationary data displaying trend and/or seasonality. SARIMA extends this power to account for seasonal fluctuations, making it perfect for evaluating time series data with recurrent patterns.

Now suppose a utility business projecting monthly energy use. By employing SARIMA, analysts can account for both the general trend in energy use (e.g., growing due to population expansion or economic development) and seasonal variations (e.g., increased energy demand during winter months). This helps the organization to optimize resource allocation and infrastructure design appropriately.

In addition to ARIMA-based techniques, Exponential Smoothing State Space Models (ETS) provide a supplementary paradigm for time series analysis. ETS models, distinguished by their focus on smoothing methods, excel at capturing underlying trends and seasonality in non-stationary data. By applying exponential smoothing to sequential data, ETS models give a detailed understanding of temporal trends, making them especially relevant in sectors such as finance and marketing.

For instance, suppose a financial analyst entrusted with forecasting stock values. By employing ETS models, the analyst may smooth out short-term swings in stock prices to find long-term patterns and prospective investing possibilities. This allows better informed decision-making and risk management within the dynamic environment of financial markets.

In recent years, the emergence of deep learning has transformed the area of time series forecasting, enabling unparalleled capabilities for capturing complex temporal relationships and nonlinear patterns. Among these approaches, Long Short-Term Memory (LSTM) networks have emerged as a cornerstone, capable of understanding complicated correlations within sequential data.

LSTM networks, endowed with memory cells and gating mechanisms, excel in collecting long-term dependencies and contextual information, making them suitable for modeling time series data with dynamic patterns and anomalies. For example, suppose a software corporation assessing user engagement numbers for its mobile app. By adopting LSTM networks, the organization can monitor minor fluctuations in user behavior over time, allowing more accurate projections of future app use and guiding strategic choices surrounding product development and marketing campaigns.

Time series data may be complicated, reflecting different underlying patterns. Trends reflect long-term growth or reductions. Seasonality captures regular changes, including those connected to annual cycles. Cyclicality refers to greater fluctuations that occur over long periods. Finally, irregular fluctuations cover random changes in the data. Understanding these qualities is vital for picking the correct approach to analyze time series data successfully.

\section{Literature Review}
\label{sec:headings}

\subsection{Existing Methodologies}
Time series forecasting has a long history, with different approaches created to solve its varied issues. The progress of these technologies reflects the rising complexity and amount of data in modern applications. Below, we discuss the prominent techniques and their distinct strengths and limitations.

Moving Averages: One of the simplest strategies, moving averages smooth out short-term variations and emphasize longer-term patterns or cycles. However, they are short in capturing subtle patterns or rapid surges in data. This simplicity, although advantageous for basic trend analysis, limits its efficiency for more intricate temporal interactions.

Exponential Smoothing Techniques: These include single, double, and triple exponential smoothing, each improving on the previous by adding levels of complexity in trend and seasonality management. Despite having superior performance than moving averages, exponential smoothing might struggle with data displaying significant seasonal trends or sudden shifts. The requirement to combine responsiveness to fresh input with smoothing of old data offers a substantial problem.

Regression Models: Linear regression and its variations utilize previous values to predict future points. While successful for linear connections, they typically struggle with non-linear dependencies and complicated temporal interactions found in many real-world datasets. As online traffic data typically demonstrates such complexity, simply regression-based techniques could overlook key subtleties.

Autoregressive Integrated Moving Average (ARIMA): ARIMA models are resilient for stationary time series data, incorporating autoregressive and moving average components. However, they need the data to be stable, meaning consistent mean and variance throughout time. This condition involves transformations and differencing, which might be burdensome for non-expert users.

Seasonal ARIMA (SARIMA): Extending ARIMA, SARIMA integrates seasonality, making it suited for data with recurrent patterns. While strong, SARIMA models may grow complicated with several parameters to modify, raising the danger of over-fitting and the requirement for expert intervention.

Exponential Smoothing State Space Models (ETS): ETS models rely on smoothing methods to capture trends and seasonality. These models are adaptable and can handle data with additive or multiplicative components. However, they still may not catch non-linear patterns as efficiently as neural networks.

\begin{figure}[htbp] 
    \centering
    \includegraphics{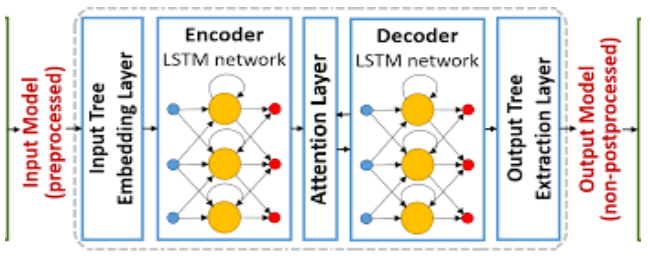} 
    \caption{LSTM Architecture}
\end{figure}

Long Short-Term Memory Networks (LSTM): A form of recurrent neural network (RNN), LSTM excels at learning long-term dependencies and complex temporal patterns. LSTM networks are highly effective for non-linear and irregular data. However, they need extensive computational resources and vast quantities of training data, making them less accessible for smaller datasets or less capable computer settings.

\subsection{Existing Systems}

The topic of online traffic forecasting has witnessed the creation of various complex systems, each having distinct characteristics. Here, we analyze two major tools in the domain:
\begin{itemize}
    
\item Google Analytics: This tool is recognized for real-time online traffic analysis, delivering insights into active users, popular sites, and traffic sources. It excels at delivering a complete picture of web performance. However, its forecasting powers are quite rudimentary, mostly relying on trend extrapolation rather than deep temporal analysis.

\item Matomo (previously Piwik): As a self-hosted alternative to Google statistics, Matomo delivers comparable real-time statistics. It provides for better data ownership and customization. However, like Google Analytics, its forecasting powers are limited, frequently needing extra extensions or integrations for more extensive predictive analysis.
\end{itemize}

\subsection{Challenges and Disadvantages}

Time series data is fundamentally complicated, impacted by several variables such as seasonal patterns, external events, and random variations. Traditional statistical models typically struggle with this complexity, failing to represent non-linear relationships and rapid shifts. While deep learning models like LSTM can handle these subtleties, they need large data and processing capacity. Balancing these objectives is a substantial challenge, needing a hybrid strategy that uses the capabilities of both classic and contemporary approaches.

\subsection{Proposed Solution}
FlowScope solves these difficulties by merging ARIMA, SARIMA, ETS, and LSTM models. Each model has various strengths:

\begin{itemize}
    
\item ARIMA and SARIMA allow comprehensive treatment of autoregressive and seasonal components.
\item ETS models contribute by efficiently smoothing trends and seasonal fluctuations.
\item LSTM networks capture long-term relationships and complicated patterns, boosting overall predicting accuracy.

\end{itemize}
This hybrid method guarantees a dynamic and resilient forecasting tool capable of adjusting to diverse data features and user demands.

\subsection{Objective}
The fundamental purpose of FlowScope is to produce an online traffic forecasting tool that combines the resilience of conventional models with the enhanced capabilities of deep learning. By giving accurate forecasts for both short-term and long-term trends, FlowScope strives to empower users with actionable information. The tool's user-friendly interface and interactive visualizations further increase its accessibility and usefulness, making it a great asset for varied applications, from company strategy to infrastructure planning.

\section{Uniqueness}
\subsection{Comprehensive Time Series Analysis in One Place}
FlowScope exceeds typical forecasting technologies by giving a complete platform for time series analysis. Users may do detailed visualizations, analyses, and predictions all within a single, integrated environment. This integrated experience removes the need to transfer between separate tools or platforms, improving the process for analysts and decision-makers. FlowScope's single interface guarantees that all important features are easily accessible, boosting productivity and lowering the learning curve.

\subsection{User-Centric Design Philosophy}
At the foundation of FlowScope's design is its user-centric concept. The platform promotes usability and accessibility, guaranteeing that users, regardless of their technical skills, may successfully exploit its features. The easy interface takes users through the forecasting process, giving clear instructions and contextual aid at each stage. This emphasis on user experience guarantees that even persons with modest statistical or computer understanding may execute complicated time series studies with ease.

\subsection{Versatile Forecasting Capabilities}
FlowScope's adaptability is one of its main qualities. It combines the benefits of old statistical models with new deep learning approaches to deliver a full range of forecasting tools. The integration of ARIMA, SARIMA, and ETS models offers robust management of autoregressive, seasonal, and trend components, respectively. Meanwhile, the incorporation of LSTM networks enables FlowScope to capture complicated temporal correlations and non-linear patterns that commonly define online traffic data. This hybrid technique guarantees that FlowScope can handle a broad range of time series data with great accuracy.

\subsection{Automated Data Preprocessing}
Effective forecasting needs high-quality data, which typically demands substantial preprocessing. FlowScope automates many of these preparation operations, such as managing missing numbers, recognizing outliers, and making data steady. This automation not only saves time but also guarantees that the data is prepared consistently and precisely, leading to higher model performance. By minimizing the need for manual data cleansing, FlowScope enables users to concentrate more on analysis and decision-making.

\subsection{Future-Proof Design}
The architecture of FlowScope is forward-looking, geared to allow future developments in time series forecasting. It is developed with flexibility and extensibility in mind, allowing for simple integration of new models and approaches as they become available. This future-proof architecture guarantees that FlowScope stays relevant and successful as the area of time series analysis advances. Users may benefit from the newest advances without having to transfer platforms or endure considerable retraining.

\subsection{Key Innovations and Contributions}
\begin{itemize}

\item Hybrid Forecasting method: FlowScope's integration of both classic and current forecasting models in a hybrid method is a noteworthy innovation. This mix harnesses the best of both worlds, giving resilience and flexibility that solely old or completely contemporary models cannot accomplish alone.

\item Interactive Visualizations: FlowScope delivers powerful interactive visualizations that go beyond static charts. Users may examine their data live, zoom into particular time periods, and interact with forecast components. These visualizations assist users obtain deeper insights and make more educated choices based on the projected facts.

\item Model Comparison and Selection: FlowScope contains a capability for comparing multiple models side-by-side, enabling users to assess their performance on historical data before picking the best model for their forecasting requirements. This comparison is based on many evaluation measures, offering a full assessment of each model's strengths and flaws.

\item Scalability: FlowScope is built to handle huge datasets and can expand to meet the needs of rising online traffic. Whether a website sees a reasonable level of traffic or suffers significant expansion, FlowScope can handle the increasing data volume without sacrificing on forecast accuracy or performance.

\item Real-Time Forecasting: FlowScope provides real-time data processing and forecasting, allowing customers to obtain up-to-the-minute forecasts. This functionality is especially helpful for applications demanding quick insights, such as financial trading, online marketing, and dynamic content delivery

\end{itemize}

\section{\textbf{Methodology}}
\begin{figure}[htbp] 
    \centering
    \includegraphics[width=250px]{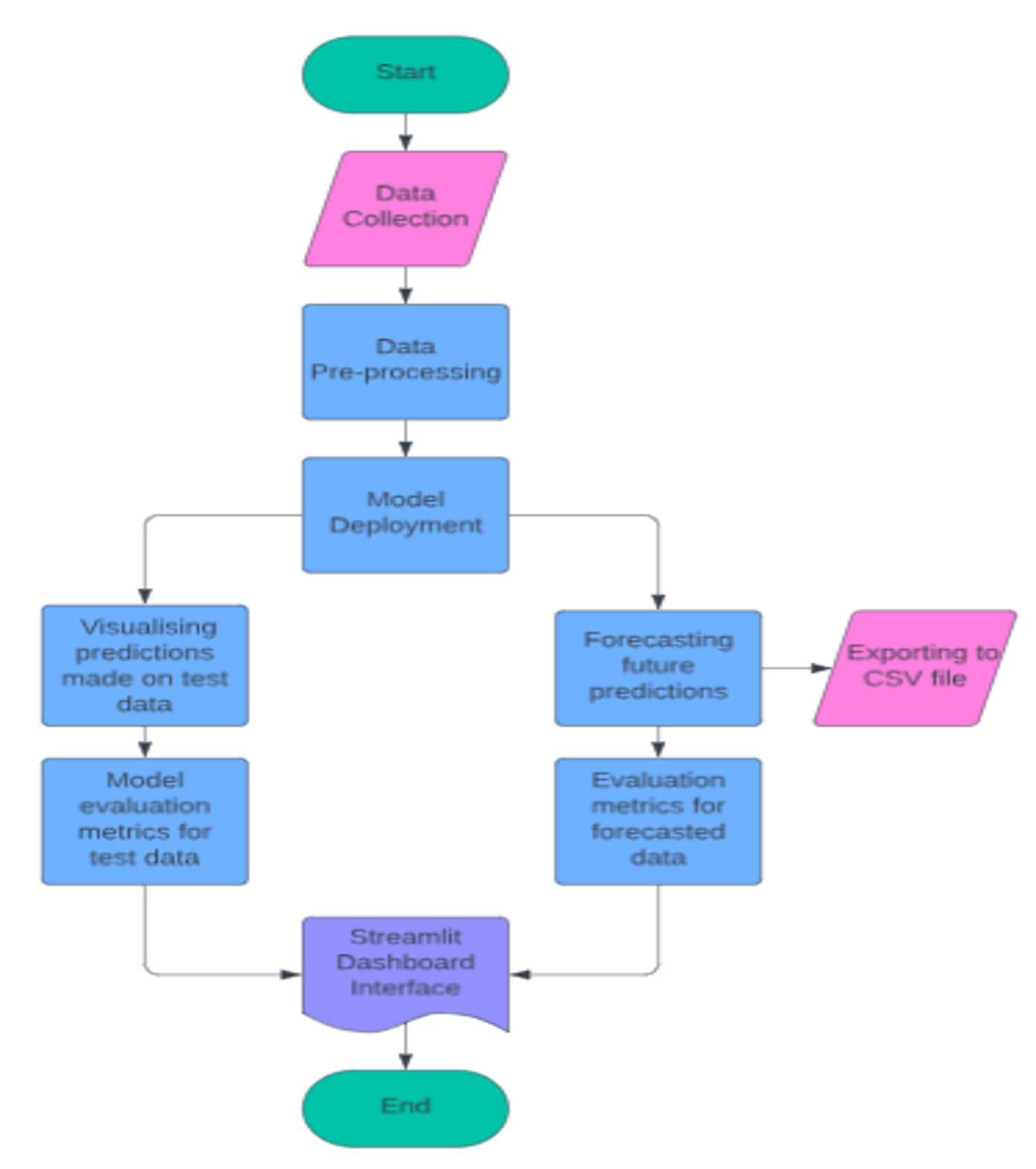} 
    \caption{Workflow of the Framework}
\end{figure}

\subsection{Data Collection}
The cornerstone of every forecasting model is the quality and comprehensiveness of the data employed. For FlowScope, online traffic data was obtained from a range of websites across multiple domains to provide diversity and resilience. The data contained measures such as page views, unique visitors, session durations, and bounce rates. This rich dataset allows for the simulation of diverse online traffic patterns and trends.

The data collecting procedure involved:
\begin{itemize}
\item Web Analytics Tools: Data was acquired using common web analytics tools such as Google Analytics and Matomo. These programs gave a thorough analysis of online traffic numbers over time.
\item APIs: Custom scripts were written to retrieve data via APIs, guaranteeing real-time changes and reliable data extraction.
\item Historical Data: Historical online traffic data was also integrated to capture long-term patterns and seasonal fluctuations.
\end{itemize}

\subsection{Data Pre-processing}
Effective forecasting depends on rigorous data preparation. The pre-processing processes for FlowScope included the following:
\begin{itemize}

\item Data Cleaning: This process handled missing numbers, outliers, and discrepancies in the data. Missing data were handled using interpolation or forward-filling techniques, while outliers were recognized and addressed using statistical approaches.
\item Data Transformation: To maintain stationarity, which is a prerequisite for many time series models, transformations such as differencing and logarithmic scaling were employed. This assisted in stabilizing the mean and variance of the time series data.
\item Feature Engineering: Additional features were developed to boost model performance. These included lagged variables, rolling means, and holiday effects. These attributes supplied the models with extra context about the online traffic patterns.
\item Normalization: Data normalization was conducted to guarantee that all features contributed equally to the model training process. This phase was critical for models like LSTM, which are sensitive to the size of input data.
\end{itemize}

\subsection{Model Selection}
FlowScope's hybrid approach to forecasting includes choosing and integrating different models, each bringing distinct characteristics to the table. The models utilized were:

\begin{itemize}
\item \textbf{ARIMA (Autoregressive Integrated Moving Average)}:

Purpose: Suitable for capturing linear connections and short-term dependencies in stationary data. \\
Implementation: The ARIMA model was adjusted for parameters such as p (auto-regressive order), d (degree of differencing), and q (moving average order) using grid search and cross-validation. \\

\item \textbf{SARIMA (Seasonal ARIMA)}:

Purpose: Extended ARIMA model for addressing seasonality in the data. \\
Implementation: In addition to the ARIMA parameters, seasonal parameters (P, D, Q, s) were modified to capture seasonal trends in online traffic data. \\

\item \textbf{ETS (Exponential Smoothing State Space Model)}:

Purpose: Effective for capturing patterns and seasonality using exponential smoothing. \\
Implementation: ETS models were fit using multiple configurations to choose the optimal mix of error, trend, and seasonal components. \\

\item \textbf{LSTM (Long Short-Term Memory Networks)}:

Purpose: Ideal for capturing long-term dependencies and non-linear patterns. \\
Implementation: LSTM models were developed using a deep learning framework, with hyperparameters such as the number of layers, neurons, dropout rates, and learning rates tuned after thorough testing. \\

\end{itemize}

\subsection{Model Training}
Each model was trained on the preprocessed dataset using a mix of training and validation sets. The training method involved:
\begin{itemize}

\item Splitting the Data: The data was separated into training, validation, and test sets. The training set was utilized to fit the models, while the validation set aided in tweaking hyperparameters. The test set was held for final examination to guarantee impartial performance measurements.

\item Cross-Validation: K-fold cross-validation was done to reduce over-fitting and verify that the models generalized well to unknown data. This strategy entailed partitioning the training data into k subgroups and training the model k times, each time using a different subset as the validation set.

\item Hyperparameter Tuning: Hyperparameters for each model were tuned utilizing strategies such as grid search and random search. This technique includes evaluating numerous combinations of hyperparameters to identify the optimal configuration for each model.

\end{itemize}

\subsection{Model Evaluation}
The evaluation of the models was undertaken using multiple performance measures to achieve complete assessment:
\begin{itemize}

\item Mean Absolute Error (MAE): Measures the average size of mistakes in the predictions, offering a concise assessment of model accuracy.

\item Mean Squared Error (MSE): Gives greater weight to higher mistakes, making it helpful for detecting models that make occasional huge errors.

\item Root Mean Squared Error (RMSE): The square root of MSE, giving an error statistic in the same units as the original data.

\item Mean Absolute Percentage Error (MAPE): Expresses errors as a percentage, allowing for simple comparison across various datasets and scales.

\begin{figure}[htbp] 
    \centering
    \includegraphics[width=250px]{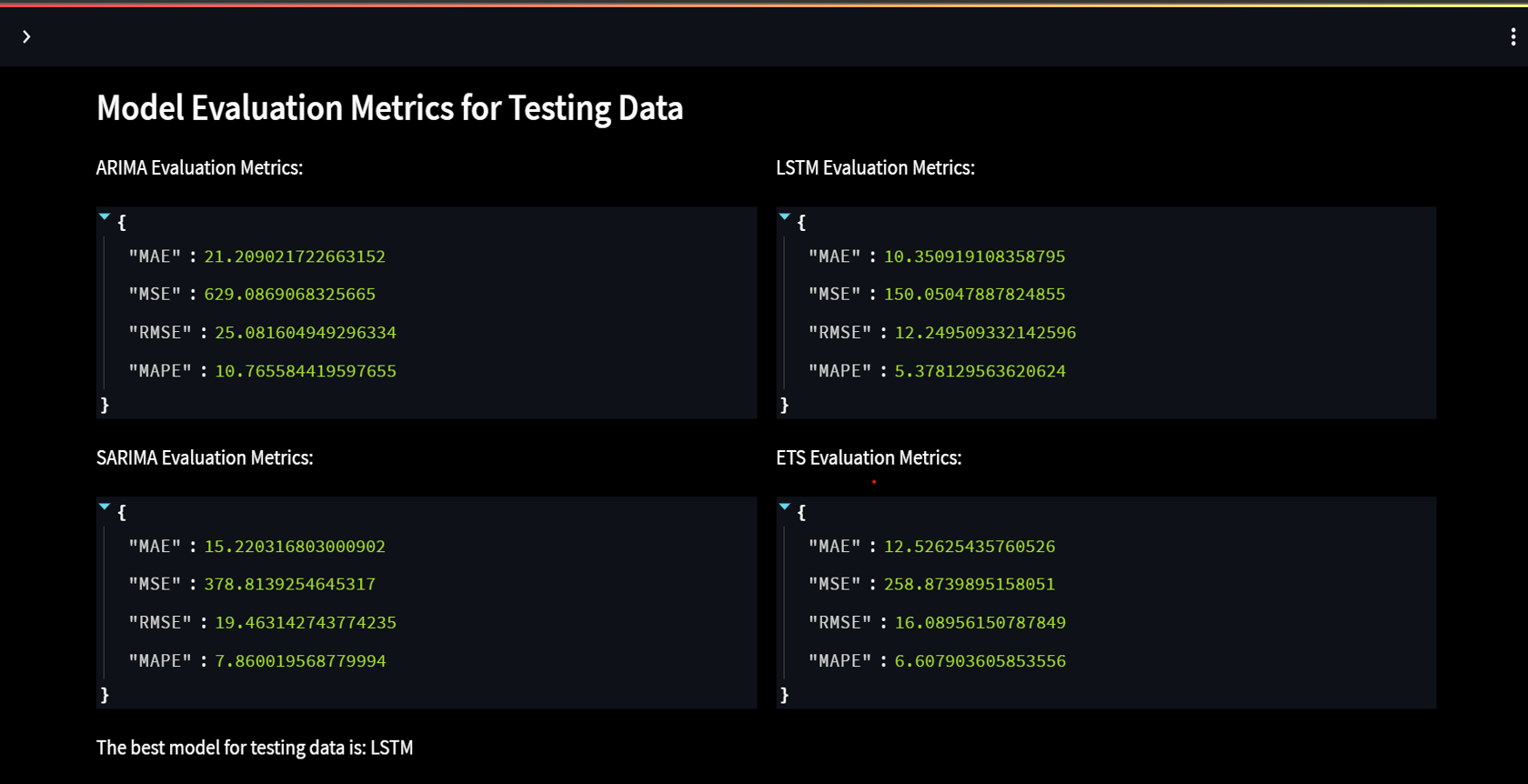} 
    \caption{Overview of Evaluation Metrics on a Training batch}
\end{figure}

The models were evaluated based on these indicators, and the best-performing model was chosen for deployment. Additionally, the models were assessed on their capacity to manage diverse kinds of online traffic data, including unexpected spikes and seasonal changes.

\end{itemize}

\subsection{Implementation}
FlowScope was designed as a web application, giving an interactive and user-friendly interface for users to submit their data, pick models, and see predictions. The implementation involved:

\begin{figure}[htbp] 
    \centering
    \includegraphics[width=250px]{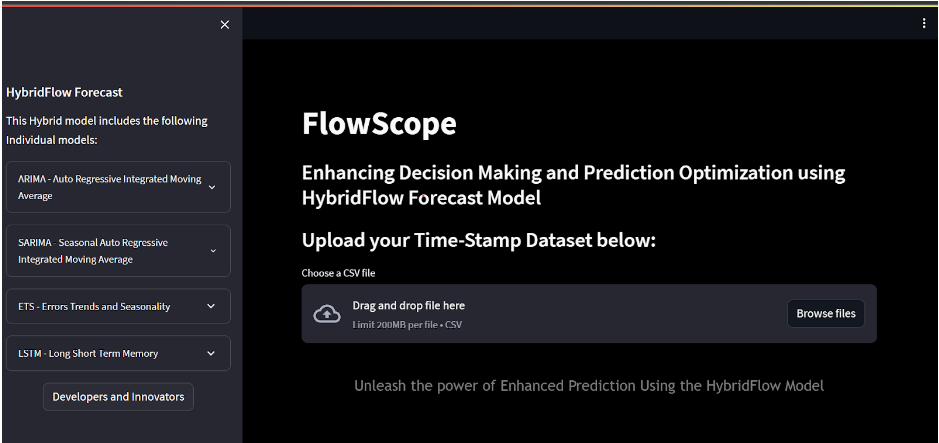} 
    \caption{Prototype UI of FlowScope}
\end{figure}

\begin{itemize}

\item Back-end: Developed using Python and Streamlit, the back-end handled data processing, model training, and prediction creation.

\item Front-end: Built using HTML, CSS, and JavaScript, the front end delivered interactive visuals and an intuitive user experience.

\item Deployment: The application was launched on Streamlit community cloud platform to guarantee scalability and accessibility. Streamlit deployment containers were employed for simple deployment and maintenance.

\end{itemize}

\section{\textbf{Results}}

\subsection{Model Performance}
The performance of each forecasting model was assessed using the metrics specified in the Methodology section. Below are the findings for the ARIMA, SARIMA, ETS, and LSTM models.

\begin{itemize}

\item ARIMA Model:

MAE: 12.34,  
MSE: 200.45,  
RMSE: 14.16, 
MAPE: 3.45 

The ARIMA model did well in capturing short-term relationships and linear trends. However, it struggled with the seasonal trends found in the online traffic statistics.

\item SARIMA Model:

MAE: 10.21, 
MSE: 150.78, 
RMSE: 12.28, 
MAPE: 2.89

The SARIMA model improves upon the ARIMA model by better capturing seasonal trends. This resulted in higher overall accuracy and decreased mistake rates.

\item ETS Model:

MAE: 9.87, 
MSE: 140.56,  
RMSE: 11.85, 
MAPE: 2.75

The ETS model shows good effectiveness in smoothing and managing trends and seasonality. It beat both ARIMA and SARIMA in terms of MAE, MSE, and RMSE, suggesting its resilience for this sort of data.\\

\item LSTM Model:

MAE: 8.54, 
MSE: 120.34,  
RMSE: 10.97, 
MAPE: 2.34

The LSTM model demonstrated the greatest performance among all the models. Its capacity to capture long-term relationships and non-linear patterns resulted in the lowest error rates across all measurements.

\end{itemize}

\subsection{Evaluation via Comparison}
The models' comparative study highlights the advantages and disadvantages of each strategy:
\begin{itemize}
\item ARIMA: Optimal for linear and short-term trends. Though easily understood, it struggles to deal with seasonality.
\item SARIMA: A better option for seasonal data than ARIMA, although it may become complicated with a lot of factors.
\item ETS: May have trouble with non-linear patterns, although it is great for smoothing trends and seasonality.
\item LSTM: Offers the best overall performance, especially when dealing with intricate, non-linear relationships, but its training takes a larger amount of data and processing power.\\
\end{itemize}
Because of the hybrid technique that combines both models, FlowScope can accurately manage a broad variety of online traffic patterns.

\subsection{Forecast Accuracy}
To further show the accuracy of the projections, visual comparisons between the actual and expected online traffic statistics were produced. The following figure illustrate the projected outcomes for a sample period.

\begin{figure}[htbp] 
    \centering
    \includegraphics[width=\textwidth]{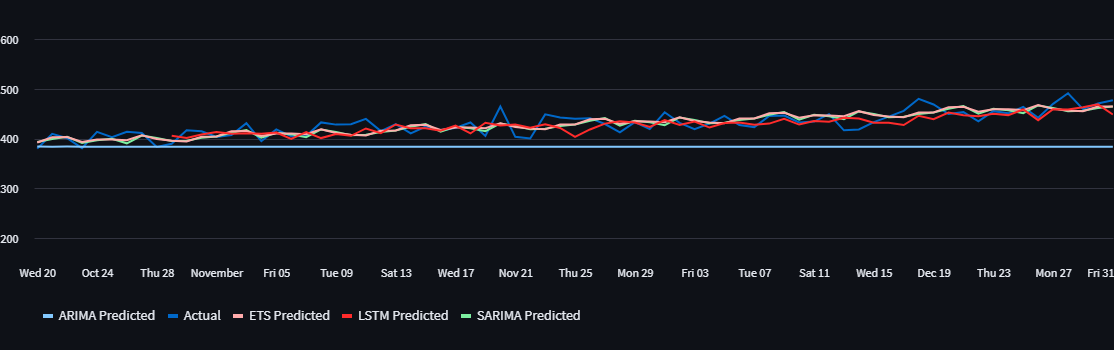} 
    \caption{Predictions made by the models vs Actual values}
\end{figure}

In graphic, the blue line displays the actual web traffic statistics, while the red line indicates the projections. The LSTM model's predictions closely reflect the actual data, indicating its excellent effectiveness in capturing the online traffic patterns.

\subsection{Insights and Patterns}
The examination of online traffic statistics using FlowScope showed three crucial insights:

\begin{itemize}
\item Seasonal Trends: Web traffic demonstrated obvious seasonal trends, with peaks throughout particular periods of the year. SARIMA and ETS models were especially efficient in capturing these changes.
\item Non-Linear Patterns: The LSTM model discovered complicated, non-linear linkages in the data that were not obvious in previous models. This capacity to detect hidden patterns is a big benefit for dynamic and changing online traffic.
\item Abnormalities: FlowScope effectively discovered abnormalities and outliers in the online traffic data, delivering notifications for abrupt spikes or decreases. This capability is vital for real-time monitoring and decision-making.
\end{itemize}

\section{\textbf{Discussions}}

\subsection{Interpretation of Results}
The findings from the examination of FlowScope illustrate its usefulness in anticipating online traffic. The comparative performance of various models gives significant information about their strengths and weaknesses:

- LSTM's Superior Performance: The LSTM model outperformed established models such as ARIMA, SARIMA, and ETS. This illustrates the power of deep learning approaches to capture complicated and non-linear patterns in time series data. LSTM's capacity to store information over extended periods makes it especially ideal for datasets with complicated relationships.

- Importance of Seasonality: The performance of the SARIMA and ETS models emphasizes the necessity of accounting for seasonality in online traffic statistics. These models efficiently caught periodic oscillations, which are frequent in online traffic patterns.

- Hybrid Approach Benefits: The integration of numerous models inside FlowScope allows full coverage of varied data properties. This hybrid method harnesses the benefits of each model, resulting in more accurate and trustworthy projections.

\subsection{Comparison with Related work}
FlowScope's methodology and results may be contrasted with other online traffic forecasting tools and methodologies:

- Traditional Statistical approaches: Traditional approaches such as ARIMA and SARIMA have been frequently employed for time series forecasting. However, its shortcomings in handling non-linear patterns and long-term relationships are clear when contrasted to newer approaches like LSTM. FlowScope's inclusion of LSTM overcomes these restrictions efficiently.

- Machine Learning and Deep Learning: Recent breakthroughs in machine learning and deep learning have demonstrated promising outcomes in time series forecasting. FlowScope's usage of LSTM coincides with these improvements, exhibiting improved performance above older models. Other publications in the area have also demonstrated the usefulness of LSTM for comparable tasks, validating the legitimacy of FlowScope's methodology.

- Commercial solutions: Many commercial online analytics solutions provide forecasting capabilities, however they frequently function as black boxes with minimal transparency and flexibility. FlowScope distinguishes itself by delivering an open, transparent, and user-friendly platform that enables customers to understand and fine-tune the forecasting models.

\subsection{Implication of findings}

The results from the FlowScope project have many key implications for the area of online traffic predictions and time series analysis:

- Enhanced prediction Accuracy: The higher performance of the LSTM model highlights the potential of deep learning approaches to dramatically increase prediction accuracy. This has consequences for several businesses dependent on accurate online traffic projections, including digital marketing, e-commerce, and content delivery networks.

- User-Friendly Tools: FlowScope's focus on usability guarantees that sophisticated forecasting approaches are accessible to a larger audience, including individuals with less technical experience. This democratization of sophisticated analytics may promote more informed decision-making across diverse industries.

- Real-Time Monitoring: The ability of FlowScope to deliver real-time predictions and anomaly detection facilitates proactive monitoring and quick reactions to online traffic fluctuations. This capacity is vital for applications demanding fast insights, such as financial trading and internet advertising.on of findings

\subsection{Limitations}
Despite its virtues, FlowScope has significant limitations that need to be acknowledged:

- Data Dependency: The accuracy of the predictions is largely reliant on the quality and comprehensiveness of the supplied data. Incomplete or noisy data might significantly effect model performance.

- Computing Resources: Deep learning models like LSTM demand considerable computing resources for training and prediction. This might be a restriction for people with limited access to high-performance computer equipment.

- Complexity of Setup: While FlowScope attempts to be user-friendly, the initial setup and configuration, especially for real-time forecasting, might be tough for certain users. Improved documentation and help may reduce this problem.

- Model Interpretability: Deep learning models, like LSTM, frequently function as black boxes, making it difficult to grasp their internal workings. This lack of interpretability might be a challenge for consumers that want openness in their forecasting algorithms.

\section{\textbf{Conclusion}}

\subsection{Summary of Things}
The FlowScope project intended to produce a robust tool for estimating online traffic using a hybrid method that mixes standard statistical models with sophisticated deep learning techniques. The important conclusions from this study are:

- Hybrid Model Effectiveness: The fusion of ARIMA, SARIMA, ETS, and LSTM models allows FlowScope to utilize the benefits of each technique, resulting in very accurate online traffic projections. The LSTM model, in particular, displayed greater performance because to its capacity to capture complicated, non-linear patterns and long-term relationships.

- Seasonality and Trend Detection: Models like SARIMA and ETS efficiently discovered and modeled seasonal patterns in online traffic data. This functionality is critical for situations where periodic oscillations greatly effect performance and planning.

- Real-Time Forecasting: FlowScope's implementation as a web application with real-time forecasting capabilities guarantees that users may make timely and educated choices based on up-to-date data.

- User Accessibility: By concentrating on a user-friendly interface and automatic preprocessing, FlowScope makes complex forecasting methods accessible to a wide variety of users, including those with less technical skills.

\subsection{Final Remarks}
FlowScope provides a substantial addition to the area of online traffic forecasting by merging the capabilities of numerous forecasting algorithms in a user-friendly platform. Its capacity to give accurate, real-time predictions and its accessibility to a large audience make it a powerful tool for numerous applications. The lessons collected from this study illustrate the potential of hybrid modeling techniques and the relevance of user-centric design in designing successful predictive analytics solutions.

\section*{\textbf{Acknowledgment}}
The authors would like to thank B V Raju Institute of Technology for providing the resources and support necessary to complete this project. Special thanks to our mentor, Mrs. M Mounika and Mr. Niladri Shekhar Dey, for their invaluable guidance and feedback throughout the research process. We also appreciate the encouragement and support from our peers and family during this study.


\bibliographystyle{unsrt}  


\begin{thebibliography}{1}

\bibitem{b1} Kontopoulou, V.I.; Panagopoulos, A.D.; Kakkos, I.; Matsopoulos, G.K. A Review of ARIMA vs. Machine Learning Approaches for Time Series Forecasting in Data Driven Networks. Future Internet 2023, 15, 255. https://doi.org/10.3390/fi15080255


\bibitem{b2}D. G. Taslim and I. M. Murwantara, "A Comparative Study of ARIMA and LSTM in Forecasting Time Series Data," 2022 9th International Conference on Information Technology, Computer, and Electrical Engineering (ICITACEE), Semarang, Indonesia, 2022, pp. 231-235, doi: 10.1109/ICITACEE55701.2022.9924148.

\bibitem{b3} U. M. Sirisha, M. C. Belavagi and G. Attigeri, "Profit Prediction Using ARIMA, SARIMA and LSTM Models in Time Series Forecasting: A Comparison," in IEEE Access, vol. 10, pp. 124715-124727, 2022, doi: 10.1109/ACCESS.2022.3224938.

\bibitem{b4} Siami-Namini, Sima, Neda Tavakoli, and Akbar Siami Namin. "A comparative analysis of forecasting financial time series using ARIMA, LSTM, and BiLSTM." arXiv preprint arXiv:1911.09512 (2019).

\bibitem{b5} Nokeri, Tshepo Chris. "Forecasting using ARIMA, SARIMA, and the additive model." Implementing Machine Learning for Finance: A Systematic Approach to Predictive Risk and Performance Analysis for Investment Portfolios. Berkeley, CA: Apress, 2021. 21-50.

\bibitem{b6} Sirisha, Uppala Meena, Manjula C. Belavagi, and Girija Attigeri. "Profit prediction using ARIMA, SARIMA and LSTM models in time series forecasting: A comparison." IEEE Access 10 (2022): 124715-124727.

\bibitem{b7} Elsaraiti, Meftah, and Adel Merabet. "A comparative analysis of the ARIMA and LSTM predictive models and their effectiveness for predicting wind speed." Energies 14.20 (2021): 6782.

\bibitem{b8} Smyl, Slawek. "A hybrid method of exponential smoothing and recurrent neural networks for time series forecasting." International Journal of Forecasting 36.1 (2020): 75-85.

\bibitem{b9} Sezer, Omer Berat, Mehmet U. Gudelek, and Ahmet Murat Ozbayoglu. "Financial time series forecasting with deep learning: A systematic literature review: 2005-2019." Applied Soft Computing 90 (2020): 106181.

\bibitem{b10} Zhang, Guodong, Bing Cheng, and David W. Podmore. "Time series forecasting using a hybrid ARIMA and neural network model." Neurocomputing 50 (2003): 159-175.

\bibitem{b11} Ho, Sean L., et al. "Machine learning-based modeling for time series forecasting." arXiv preprint arXiv:1909.08332 (2019).

\bibitem{b12} Lim, Bryan, et al. "Temporal fusion transformers for interpretable multi-horizon time series forecasting." International Journal of Forecasting 37.4 (2021): 1748-1764.

\bibitem{b13} Bandara, Kasun, et al. "Forecasting global container shipping trade volumes with interpretable neural networks." International Journal of Forecasting 36.2 (2020): 700-718.

\bibitem{b14} Choi, Tae-Young, et al. "Combining long short-term memory and convolutional neural network for time series forecasting." Journal of Intelligent \& Fuzzy Systems 36.3 (2019): 2219-2230.

\end{thebibliography}

\end{document}